\documentclass[letterpaper, 10 pt, conference]{ieeeconf}  

\IEEEoverridecommandlockouts                              

\usepackage{graphicx} 
\usepackage{xcolor}
\title{\LARGE \bf
Designing a Socially Assistive Robot for Long-Term In-Home Use\\ for Children with Autism Spectrum Disorders
}

\author{Roxanna Pakkar, Caitlyn Clabaugh, Rhianna Lee, Eric Deng, and Maja J Matari\'c
\thanks{Roxanna Pakkar, Caitlyn Clabaugh, Rhianna Lee, Eric C. Deng, and Maja J. Matari\'c are with the Interaction Lab, Department of Computer Science, University of Southern California, Los Angeles, CA 90089, USA. {\tt\small {pakkar@usc.edu, clabaugh@usc.edu, rdorothyl@gmail.com, ericcdeng@gmail.com, mataric@usc.edu}}}%
}

\begin{document}

\maketitle

\thispagestyle{empty}
\pagestyle{empty}

\begin{abstract}

Socially assistive robotics (SAR) research has shown great potential for supplementing and augmenting therapy for children with autism spectrum disorders (ASD). However, the vast majority of SAR research has been limited to short-term studies in highly controlled environments. The design and development of a SAR system capable of interacting autonomously {\it in situ} for long periods of time involves many engineering and computing challenges. This paper presents the design of a fully autonomous SAR system for long-term, in-home use with children with ASD. We address design decisions based on robustness and adaptability needs, discuss the development of the robot's character and interactions, and provide insights from the month-long, in-home data collections with children with ASD. This work contributes to a larger research program that is exploring how SAR can be used for enhancing the social and cognitive development of children with ASD.

\end{abstract}

\section{INTRODUCTION}

The Centers for Disease Control estimate that autism spectrum disorders (ASD) affect 1 out of 59 children in the United States \cite{christensen2016prevalence}. ASD is characterized by a variety of developmental delays and difficulties in communication and social skills, ranging in their severity and manifestations \cite{kanne2008diagnostic}. Early therapeutic and educational interventions are critical for individuals with ASD \cite{dawson2008early} \cite{ospina2008behavioural}, but the  costs of those services, along with the associated transportation and parent/caregiver time, make them inaccessible to many \cite{lavelle2014economic}. Advances in computing and robotics have provided means of supplementing such health services. Research has shown that in human-robot interaction (HRI) the physical embodiment of the support agent can increase user adherence \cite{bainbridge2008effect}, social engagement \cite{wainer2006role} \cite{lee2006physically}, and cognitive learning gains \cite{leyzberg2012physical}, among other desired outcomes. Additionally, clinical research with robots has demonstrated that they stimulate mirror neurons \cite{gazzola2007anthropomorphic}, which are critical for theory of mind \cite{gallese1998mirror} and motor skills \cite{fridin2014embodied}.

The field of socially assistive robotics (SAR) aims to develop robot systems that support human health and wellness through co-present, social interaction \cite{mataric2016socially}. SAR research has demonstrated the unique social capacity of robots to support children in learning \cite{clabaugh2019escaping}. In ASD in particular, SAR has been shown to effectively mediate joint attention \cite{robins2008behaviour} and turn-taking \cite{baxter2013emergence}, increase interaction complexity \cite{dautenhahn1999robots}, and elicit social interaction \cite{robins2005robotic}. The embodied, social, and judgment-free nature of SAR makes it an effective tool to assist children with ASD in understanding and interacting socially with the world around them \cite{scassellati2012robots}.

In spite of this potential, due to the high development and research costs of long-term and real-world SAR studies, the vast majority of SAR research with children with ASD to date has been limited to short-term studies involving highly controlled interactions \cite{scassellati2012robots} \cite{clabaugh2019escaping}. While those studies have demonstrated the therapeutic potential of SAR, longer-term interventions have been shown to increase learning gains and human-robot attachment over time \cite{leite2013social} \cite{clabaugh2019escaping}. Most notably, a recent study \cite{scassellati2018improving} showed that a fully autonomous SAR system can sustain engagement, improve joint attention skills, and increase communication in children with ASD.

This paper is part of a larger, ongoing research effort to study long-term, autonomous SAR interventions. Here we focus on the system design while other works, such as \cite{ilab_iser}, provide deeper analyses of the user responses and models based on the use of the system. We present the physical setup, the robot character Kiwi, and the interaction design of the complete SAR system deployed in 6 homes of 8 children with ASD for a month each. The system design is evaluated by the development of the child-robot relationship as well as by the system's overall likability and usefulness, as reported by the participating families. 

\section{RELATED WORK}
A great deal of work has explored robotics for ASD, as reviewed by Scassellati et al. \cite{scassellati2012robots}, Begum et al. \cite{begum2016autism}, and Diehl et al. \cite{diehl2012autism}. The majority of past research has made use of commercially available robot platforms, such as Kaspar, Keepon, and the Nao, to demonstrate the viability of SAR as a therapeutic tool for children with ASD \cite{2017agencyASD}. The most common robotic platforms used have been tabletop robots, ranging in size from 10 inches (Keepon) to 2 feet (Jibo), and in form from abstract (Keepon, Jibo) to humanoid (Kaspar, Nao).  Though some past work has employed mobile robots to study topics like social space with children with ASD  \cite{feilseifercomputational}, tabletop robots have garnered the most appeal because they present fewer technical challenges and safety concerns for real-world use.  

One such platform, Jibo, was a commercially available tabletop robot with a unique, expressive swiveling motion. Scassellati et al. used Jibo in a month-long deployment with children with ASD, similar to our own \cite{scassellati2018improving}. In the study, children with ASD interacted with Jibo in their homes for 30 minutes every day, for 30 days. Over the intervention, the child participants showed improvements across a variety of social skills, especially joint attention.

While these results are promising, commercially developed robots used in SAR research, like Jibo, were not designed for use with or by children with ASD; to date, such platforms lack various features needed to meet the specific needs of long-term ASD interventions. Currently, there are no low-cost, general-purpose, adjustable autonomy systems for child-robot interaction research, especially for use in ASD. That is the motivation for the system we present: it was fully designed for long-term in-home deployments with children with ASD, involving input from the user community, and leveraging the demonstrated appeal, utility, and safety of squash-and-stretch tabletop robots.

\section{SYSTEM DESIGN}

\begin{figure}[h]
    \centering
    \includegraphics[width=3.0in]{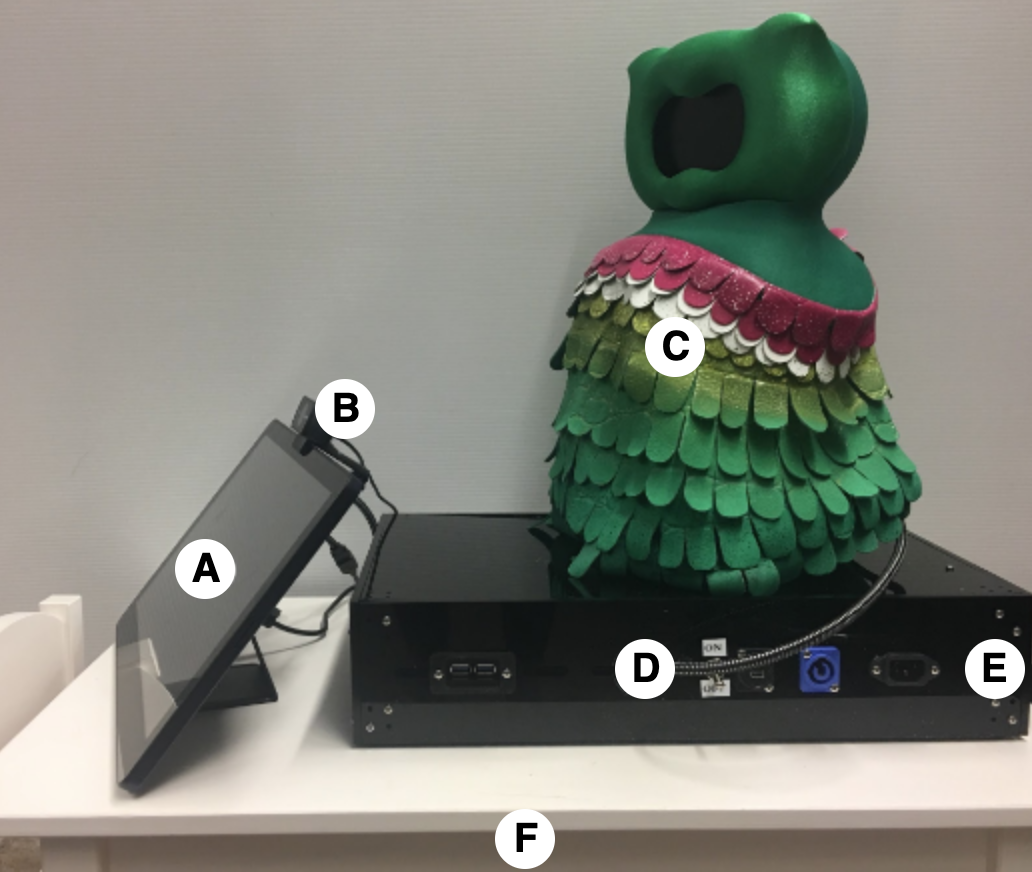}
    \caption{The physical in-home setup included a touchscreen monitor (A), a securely mounted camera (B), a SPRITE as the Kiwi character (C), an easy-to-access power switch (D), and a mini desktop computer disconnected from the Internet (E) on a child-sized table (F).}
    \label{setup}
\end{figure}

\subsection{Design Goals}
Long-term SAR interventions are challenging from a systems engineering perspective, as the deployed systems must be sufficiently robust, intuitive, and engaging to support repetitive, daily use. The following sections present the physical SAR system, robot character, and interaction design we developed and evaluated.

\subsection{Physical System}
\begin{figure}[h]
    \centering
    \includegraphics[scale = 0.4]{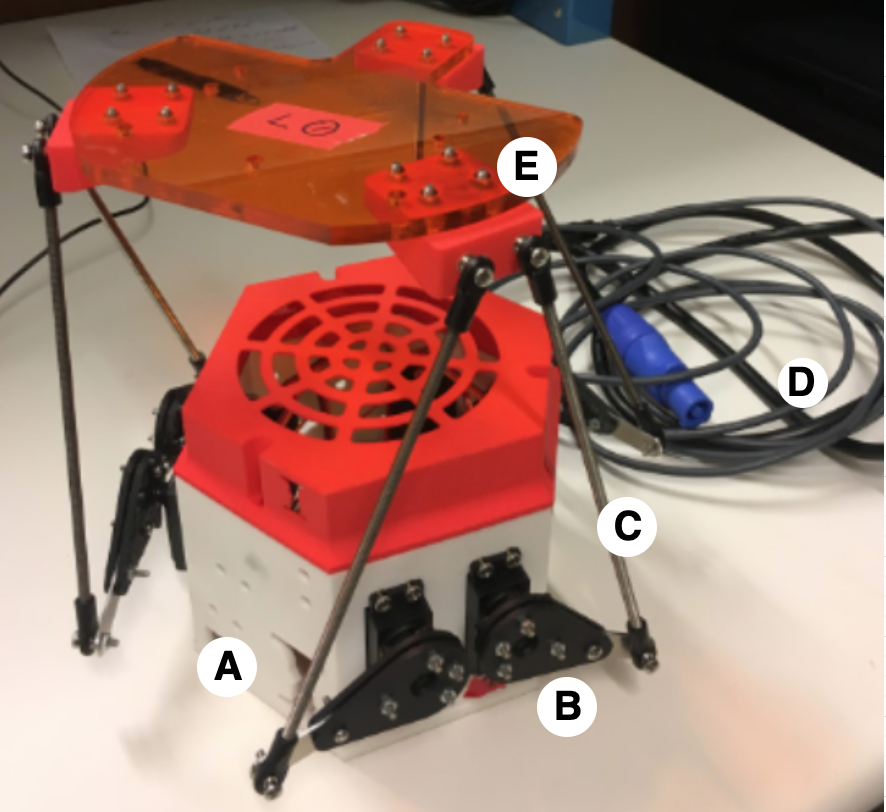}
    \caption{SPRITE's updated mechanical design consisted of: 3D printed enclosure for servo controller and electrical wiring (A); Pololu HD high-torque servos (B); threaded rods actuated to move the platform with six degrees of freedom (C); connector for robot to power supply (D); and laser-cut platform (E).}
    \label{sprite}
\label{system}
\end{figure}
\begin{figure}[h]
    \centering
    \includegraphics[scale = 0.5]{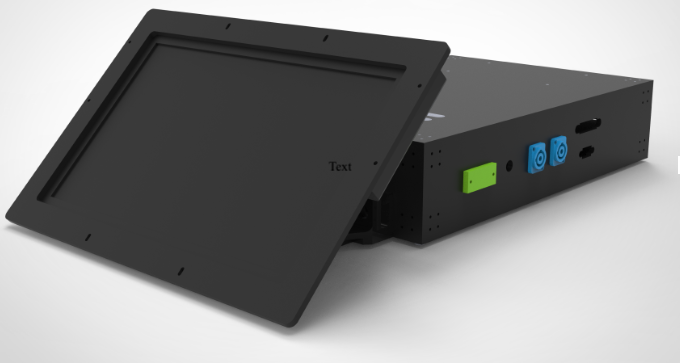}
    \caption{The display and the tabletop unit designed to safely house the SPRITE power supply, computer, speakers, and a small front-facing camera for recording audio-visual study data.}
    \label{power}
\label{system}
\end{figure}

The SAR system, shown in Figure \ref{setup}, used an updated design of the Stewart Platform Robot for Interactive Tabletop Engagement (SPRITE) \cite{short2017sprite}, shown in Figure \ref{sprite}, originally developed in our lab for ongoing SAR research. This cost effective, open-source platform consists of a 3D-printed hexagonal base that housed six servo motors controlled by one micro-controller. The motors actuated six aluminum rods that supported and moved a top plate to which a variety of 3D-printed brackets were secured. The top bracket could hold a smart phone or, as in this research, a small display, used to display the robot's face. The platform moved the robot's upper body and face using six degrees of freedom that produced the desired squash-and-stretch appearance. SPRITE can be outfitted with different exterior skins, allowing it to take on different characters depending on the research needs.

When designing systems intended for long-term use in homes with children, safety is the paramount consideration. For increased safety as well as power conservation, we replaced the ultra high torque servo motors in the original SPRITE with safer, less costly, lower power servos that supplied less torque (17 kg·cm at 6 V compared to 40 kg·cm at 7.4 V from the original design). We adjusted the 3D-printed base accordingly to fit the new servos. We also replaced the smart phone that served as the robot's face display with a lower-cost and lower-power LCD display. The final height of the SPRITE used in the in-home deployments was approximately 30cm. Collectively, these design changes lowered the robot’s overall component cost by one third, from approximately US\$1,500 to US\$1,000.

As another safety measure, a key design goal was to develop a fully enclosed system that housed the small power supply (supplying the lower-torque motors), computer, speakers, touchscreen monitor, and a small front-facing camera for recording audio-visual data, as seen in Figure~\ref{power}. The box was powered by one external AC power cord plugged into a standard wall outlet. The system was designed to be powered on and off with one easy-to-access switch and to be used without an Internet connection, to avoid network safety and accessibility concerns. 

The resulting system was a single cohesive unit that was safe, accessible, affordable, and easy to set up and use. The research team assisted child participants through an initial tutorial to ensure that they were comfortable and familiar with the system's touch-screen monitor interface. Families could turn the SAR system on and off as desired.

\subsection{Robot Character}
\begin{figure}[h]
    \centering
    \includegraphics[width = 5cm]{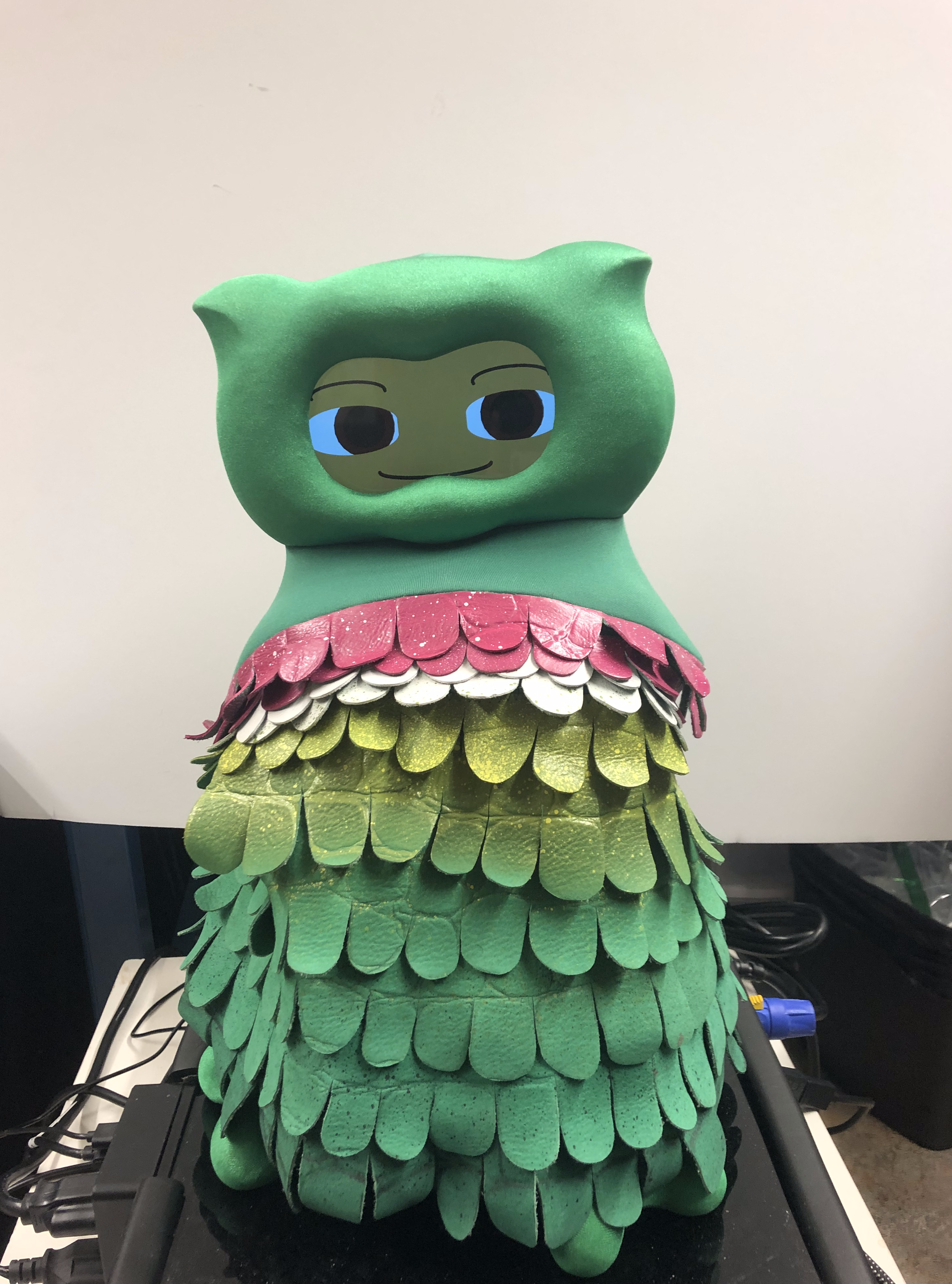}
    \caption{SPRITE as the Kiwi character}
    \label{kiwi}
\end{figure}

The robot character, Kiwi, shown in Figure \ref{kiwi}, was designed for participants with the following criteria: 
\begin{enumerate}
    \item Child user age between 3.5 years and 8 years old
    \item Stable physical, sensory (i.e., hearing and vision), and medical health
    \item English as a primary language spoken by the family
    \item Clinical diagnosis of ASD in mild to moderate ranges as described in the Diagnostic and Statistical Manual of Mental Disorders-5 \cite{van1992comparison} \cite{kan2008autism} \cite{kanne2008diagnostic} \cite{dover2007diagnose}
\end{enumerate}

Kiwi's owl-like ``skin'' and the associated character were designed to be non-threatening, gender-neutral, and simple in physical expression. The smartphone displayed an animated face with two eyes, eyebrows, and a mouth. The mouth moved using visemes; the affective facial expressions were based on the Facial Action Units Coding System (FACS) \cite{ekman1997face}.  Kiwi's speech and facial expressions were accompanied by simple squash-and-stretch and bending body motions to make communication appear natural.  As in the previous SPRITE deployments, the robot was controlled by CoRDial, the Co-Robot Dialogue system \cite{short2017sprite}, a software stack that manages the robot's speech, motor controllers, and facial animations.

Kiwi's character was designed to be a near-peer robot space explorer. To facilitate relationship-building, Kiwi used friendly informal language to relate to the child user. Kiwi introduces itself to the child as being a robot space explorer needing help to return to its home planet, and asked the child for help with a variety of game-like tasks, displayed on the touchscreen monitor. 

\subsection{Child-Robot Interaction Design}
\begin{figure}
    \centering
    \includegraphics[width=\linewidth]{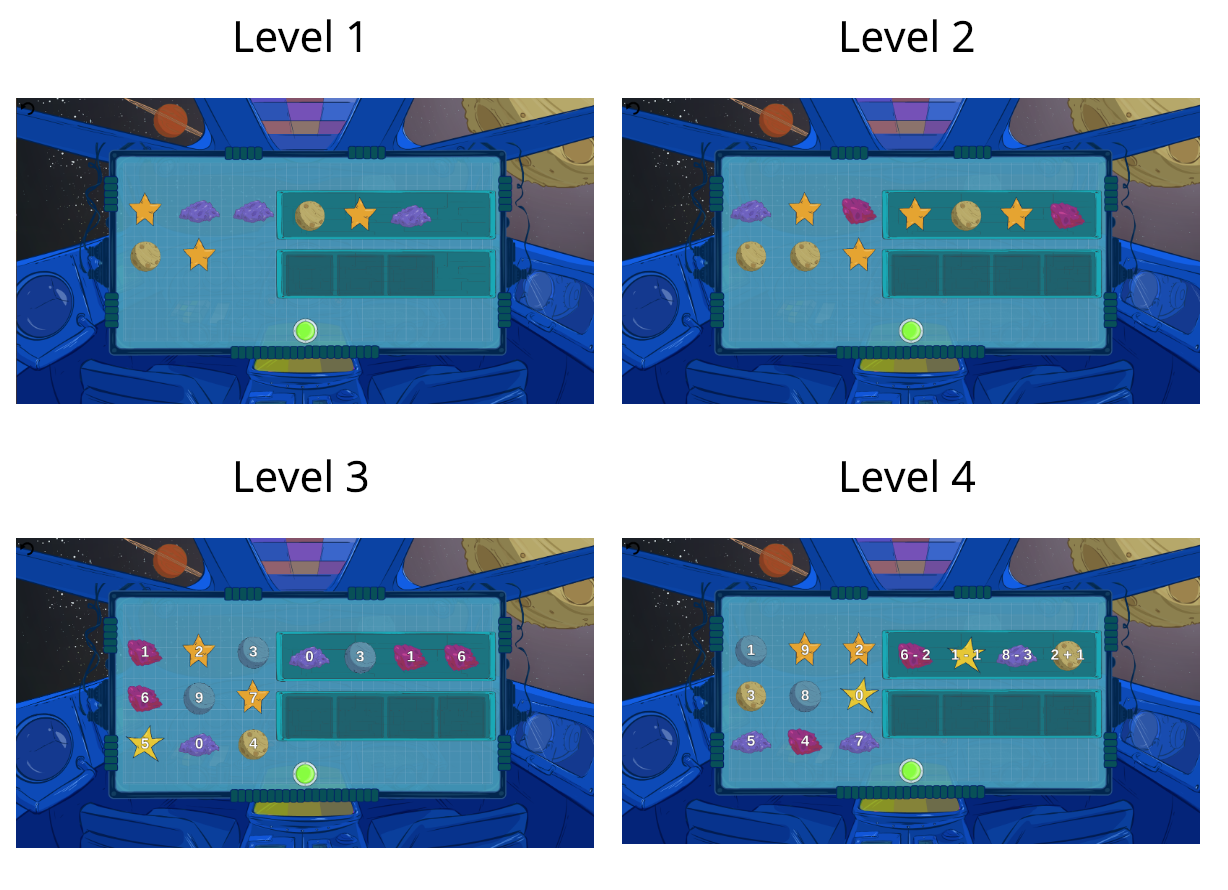}
    \caption{The child-robot interaction was designed around Kiwi as a robot space explorer. To help Kiwi return to its home planet, the child played the accompanying space-themed educational games that were displayed on a tabletop touchscreen monitor placed in front of Kiwi.}
    \label{games}
\end{figure}

\begin{figure}
    \centering
    \includegraphics[scale = 0.4]{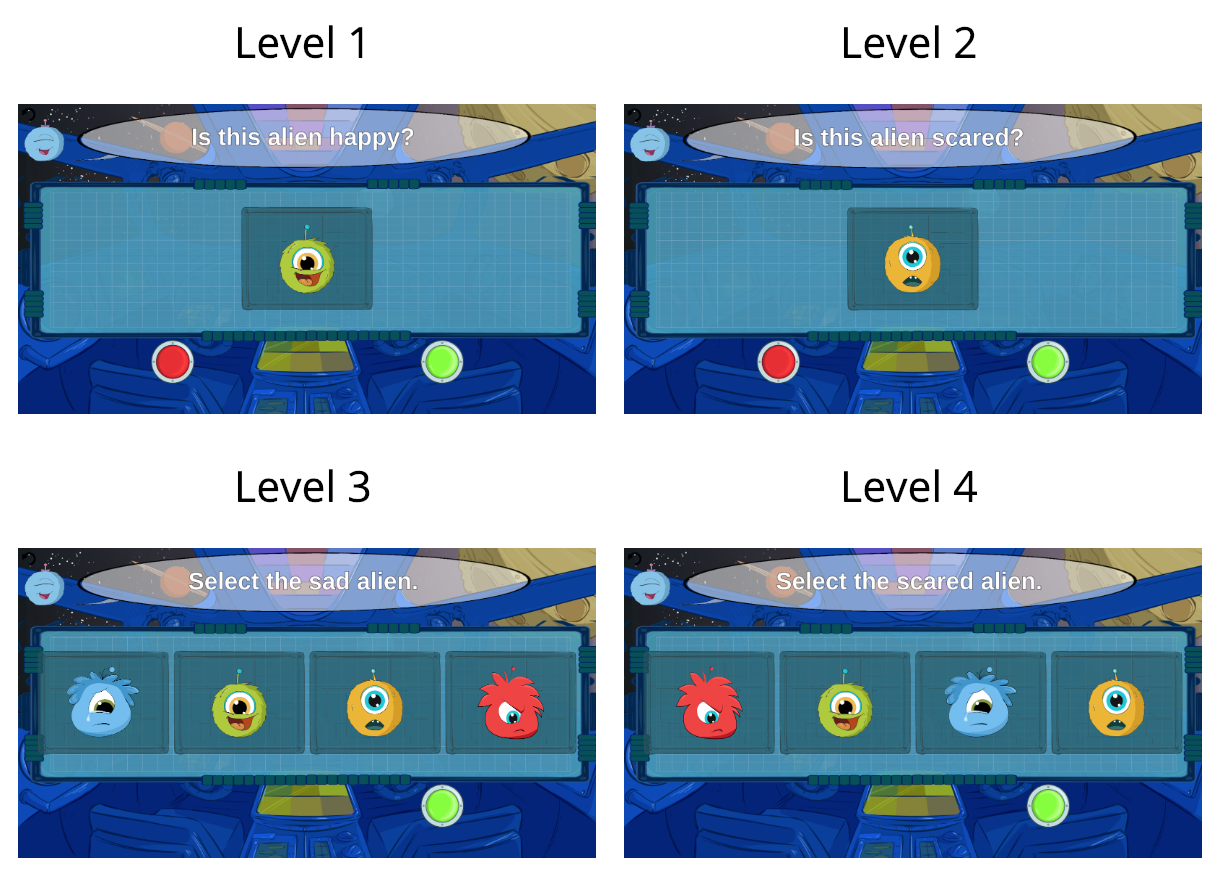}
    \caption{Example screens from social development games where participants were asked to identify emotions on the alien faces.}
    \label{emotion-game}
\end{figure}

The SAR system was designed to engage children in daily educational games depicted in Figure~\ref{games}. The games were iteratively developed concurrently with the Kiwi character in order to synergistically encourage child learning through play \cite{singer2006play} \cite{short2014train} and meet developmentally appropriate social and educational learning outcomes for children with ASD. The 10 games included in the study were designed in collaboration with education experts, teachers, and preschool-aged children \cite{clabaugh2015designing}. The games used a space theme to present and test number concepts. Activities involved sorting moon rocks and stars with the purpose of helping Kiwi fix its spaceship. Additional activities designed to support child social development were dispersed throughout the interaction (Figure~\ref{emotion-game}). These activities asked the child to identify emotions on an alien character's face. The educational games ranged in difficulty to suit the needs of each child, while the Kiwi character itself remained consistent throughout the study. As the child engaged in solving the educational games on the screen, Kiwi provided positive support, giving praise in response to correct answers (e.g., "Great job!"), and providing constructive comments in response to incorrect answers (e.g., "That's not quite right, let's try again."). 

At the start of the intervention period, Kiwi was presented to each child participant as a temporary addition to the household. This was reinforced by Kiwi's introductory dialogue in which it stated it needed help going back home.  These design decisions were made to support the temporary nature of the in-home robot deployment.

\section{EXPERIMENT DESIGN}
As part of our larger, ongoing research effort involving month-long in-home SAR deployments, this paper evaluates the described SAR system in terms of its likeability, usefulness, and relationship to the child over the 30-day deployment in six family homes of eight children with ASD.  Because of the numerous challenges of working with families with ASD in the home, including the large variance and severity of ASD symptoms across participants, and the large variance in home environments and family situations, this work follows a single-subject design \cite{horner2005use} in that child participants serve as their own respective baselines. In this paper, we focus on the insights gathered through weekly semi-structured interviews conducted with the eight participating children and their families. The interviews were used to assess robot likability, usefulness, and the child-robot relationship. Our other work reports on objective outcomes measures based on eye gaze, child vocalizations, etc.  We did not use any wearable data collection systems (e.g., EEG, GSR) because they are poorly tolerated by individuals with ASD and affect natural daily in-home activity.

Questions about Kiwi's likeability include \cite{heerink2009measuring}\cite{moon2001extending}:
\begin{itemize}
    \item Do you or your child like Kiwi? Why or why not?
    \item Do you think Kiwi makes your child happy?
    \item Does Kiwi engage your child socially, academically, and/or creatively?
\end{itemize}

\noindent Questions about Kiwi's usefulness include \cite{moon2001extending}:
\begin{itemize}
    \item Does Kiwi help your child do better on the tasks? Why or why not?
    \item How could Kiwi be more useful?
    \item How involved do you have to be while your child is playing with Kiwi?
\end{itemize}

\noindent Questions about the child-robot relationship included \cite{rau2009effects} \cite{lee2005can}:
\begin{itemize}
    \item Do you think Kiwi is your friend?
    \item Do you think Kiwi listens to you?
    \item Do you feel like Kiwi knows you?
\end{itemize}

\section{FINDINGS}

\subsection{Likeability}
\emph{How likable did the child participants and their families find Kiwi?}
\begin{figure}
    \centering
    \includegraphics[scale = 0.25]{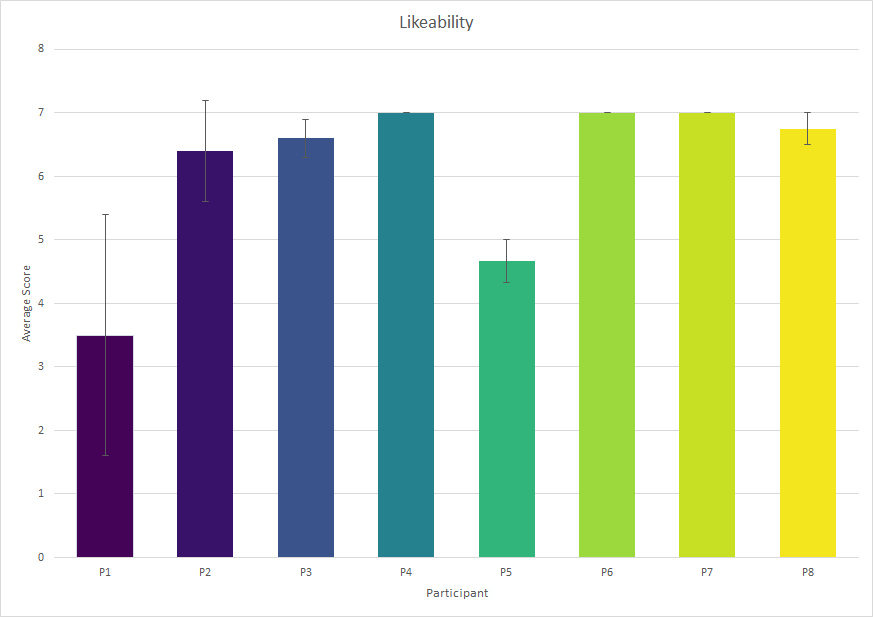}
    \caption{Average reported likeability of Kiwi rated on a five-point Likert-type scale and averaged over the intervention for each participant (mean = 6.1, SD = 1.2)}
    \label{like}
\end{figure}

As shown in Figure \ref{like}, all of the participants and their families, except for Participant 1 (P1), reported high likeability for Kiwi. Overall, P1 was an outlier in the study; across all three measures, P1 reported generally negative but highly variable ratings of the SAR system. The family stated that while he was generally comfortable with technology, he was not with the robot's physical movement and speech, ``(P1) has other apps that he uses, and because we thought this would be similar, we thought that he would be engaged...but I think that the biggest difference is the robot and I think he's scared of it.'' For a couple of weeks, the family covered Kiwi with a towel and reported that P1 had more comfortable interactions with the rest of the system. However, when uncovered again, the parents stated that P1 continued to react negatively toward the Kiwi and requested that it be removed from the home before the end of the study period.

Participant 2 (P2) reported liking Kiwi and actively talked to people about it, telling them that Kiwi is her friend. P2 said that despite Kiwi's help, the games were hard for her, and sometimes she needed motivation to continue playing. Towards the end of the study period, she still liked Kiwi but was growing tired of it and said she would not be sad when Kiwi left.

Sibling Participants 3 and 4 (P3 and P4) both reported that they liked Kiwi and grew more comfortable with it through the study period.  When asked what he tells his friends about Kiwi, P4 said ``It's awesome". P3 and P4 liked the Kiwi skin and the space theme of the games as it was something they were familiar with from school. P4 was younger than P3 and was not able to engage with the games as well as P3. Despite that, he said he liked the Kiwi character. 

Another sibling pair, Participants 5 and 6 (P5 and P6) reported generally positive interactions with Kiwi that again showed the likeability and adaptability of the system similarly to P3 and P4.  P6, the older sibling, was reported to have enjoyed every aspect of Kiwi, while it was reported that P5 took a little longer to grow accustomed to the system, but grew to like Kiwi over the month intervention.

In terms of system design, Participant 7 (P7) stated that she liked the space theme of the games as they related to other activities she did outside of the home. Participant 8 (P8) also gave overwhelmingly positive feedback about the likability of the system.

\subsection{Usefulness}
\emph{How useful did the child participants and their families find Kiwi?}
\begin{figure}
    \centering
    \includegraphics[scale = 0.25]{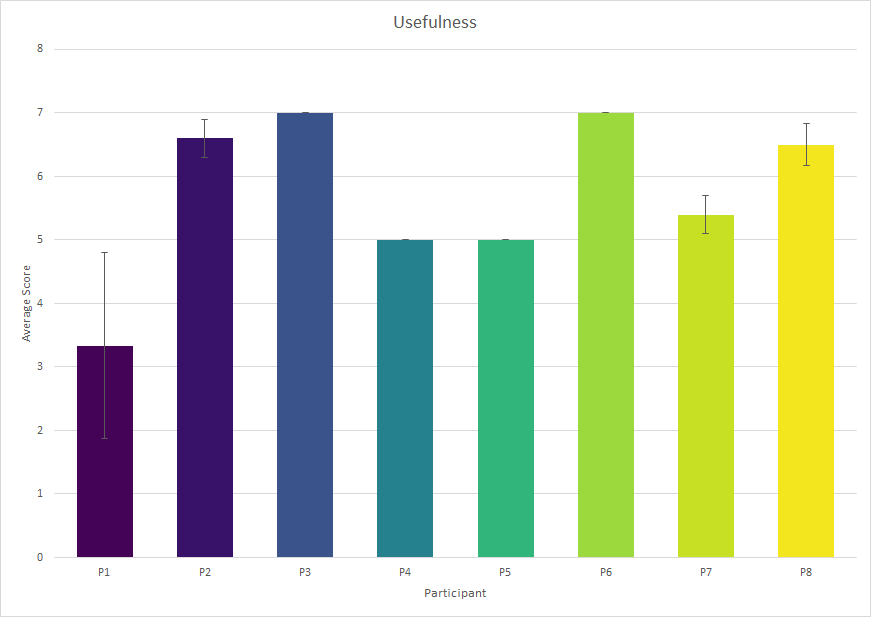}
    \caption{Average reported usefulness of Kiwi rated on a five-point Likert-type scale and averaged over the intervention for each participant (mean = 5.7, SD = 1.2)}
    \label{use}
\end{figure}

As follows from the findings for likability, P1 reported generally low usefulness ratings for the robot. Apart from the weeks during which his parents covered Kiwi, P1 did not use the system very often as he was too afraid to interact with the robot. 

The other participants had much more positive feedback on the usefulness of the system, as shown in Figure \ref{use}. P2 stated that Kiwi was helpful when she got stuck during games. Her family noted that through the interaction with the system, P2 grew more comfortable with math, which had been a difficult subject for her. Similarly, P3 reported positive experiences with both the games and Kiwi, while P4 stated that he liked Kiwi but the games were too difficult for his age level. P3 said that working with Kiwi improved his math skills. 

P5, P6, P7, and P8 also found the system to be useful. For P7, the family noted greater evidence of empathy in P7's behavior toward her peers and toward Kiwi as the interaction progressed over the study period. P8 stands out, in particular; his family noted that he learned a great deal from the system, despite initially being afraid of Kiwi's appearance. During the second week of intervention, P8's parent and sibling expressed concern that P8 avoided the more challenging problems such as subtraction. By the last week of the study period, however, P8 started to count on his fingers and was self-motivated to initiate and complete previously challenging math problems with Kiwi. This finding is further highlighted by that fact that P8 was not currently learning math at school. Later in the study, P8 reported that he had an increased enjoyment in completing his school homework, although this may or may not be related to the SAR interactions. P8's parents also noted that he demonstrated social growth over the course of the intervention; P8 became interested in astronomy, related to Kiwi's backstory, and specifically requested library books on that topic.

\subsection{Child-Robot Relationship}
\emph{How did child participants' relationships with Kiwi change over the study period?}
\begin{figure}
    \centering
    \includegraphics[scale = 0.25]{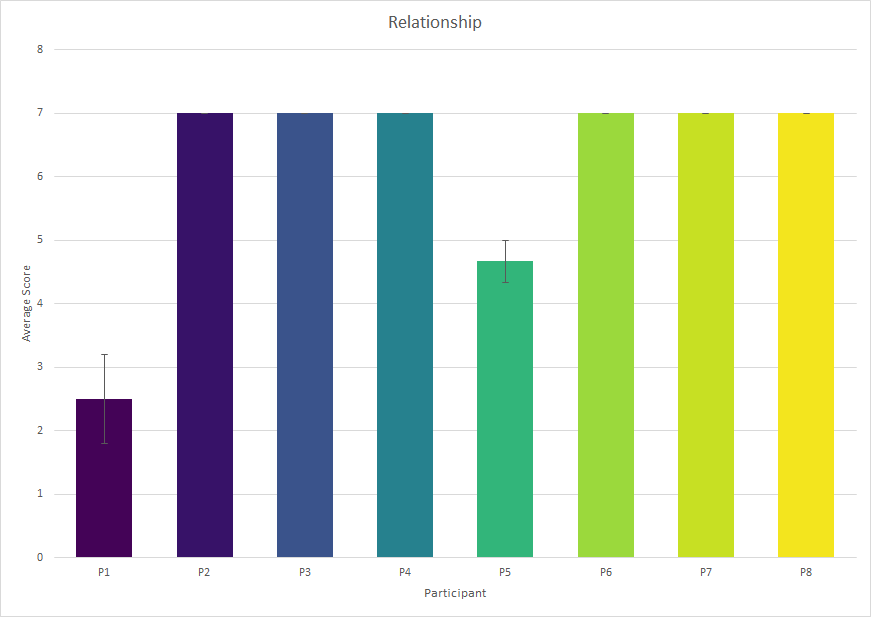}
    \caption{The closeness of the child-robot relationship was rated using a five-point Likert-type scale, averaged over all weekly interviews conducted during the intervention (mean = 6.1, SD = 1.6)}
    \label{relationship}
\end{figure}

P2-P8 reported overall positive and comfortable relationships with Kiwi, while P1 remained an outlier. When asked about the system, P1 stated ``No Kiwi,'' demonstrating his aversion to the system and to forming a relationship with the Kiwi character. 

As the interaction progressed, P2 described Kiwi as a friend and stated that the friendship was mutual, and that she thought that Kiwi liked her as well. In this way, P2 showed that Kiwi could be personified as a friend and seen as an entity separate from the games. While P2 stated that she sometimes did not like the games and found them difficult, she consistently reported positive feelings about Kiwi and the usefulness of the system.
 
P3 stated that ``Kiwi is the best'' and ``Kiwi does not make me feel bad when I get problems wrong.'' P3 and P4 grew more confident throughout the interaction. Toward the end of the study period, P3 and P4 relayed that they were worried about Kiwi leaving, and P3 said he was going to miss Kiwi. P5 and P6 had a similar response to Kiwi and also said that they thought of Kiwi as a friend. 
 
P7 did not immediately like Kiwi, and was initially reluctant to play with it. Her family reported that they had to work to persuade her to use the system and play through the games. However, after two weeks, she began to avidly seek out playing with Kiwi and ask for it as soon as she woke up in the morning. She also played with Kiwi independently, without her family in the room. Her parents also reported that P7 wanted to talk about Kiwi in school, which increased her social communication and interaction with other students, whereas previously she mainly talked with adults. She also demonstrated concern for Kiwi towards the end of the study period when she knew Kiwi was going to leave. It is important to note that, at least in this example, the participant's first impression (or first two weeks of impressions, for that matter) was not predictive of her feelings about the robot later on in the study.
 
P8 also described Kiwi as a friend, actively showing it to visitors, and going into great detail when describing his day with Kiwi. He also told Kiwi ``We make a great team'' and wanted to name a family pet after Kiwi. P8's parent stated that P8 had not had any friends at school until he started talking to children (as well as adults) more as a direct result of interactions with Kiwi.

\section{DISCUSSION}

Despite the developmental and familial differences among P2 through P8, all of these users had positive experiences with Kiwi throughout the month-long interaction. The participants' sustained interest in the system over the course of a long-term interaction demonstrates that the system's appeal surpasses the novelty effect of a new toy or playmate. P1 as the outlier shows the need for further personalization of the SAR intervention. The majority of the participants thought of Kiwi as a friend and were able to personify it as such, as evidenced by the way they talked about Kiwi to others, supporting the likability of the system. P3's family even gave design suggestions to improve the usefulness for their children, such as the addition of activities practicing language arts skills, and incorporating confirmation sounds when the participant answers a question correctly.

Throughout the study, a few technical issues arose that affected the overall scores of likability, usefulness, and relationship building over time. For P3 and P4, a software error caused Kiwi's displayed eyes to stay open for long periods. The child participants also used a water gun on that robot, but fortunately did not disable it in the process, as a result of the safety features that were part of the robot's physical design. For P5 and P6, the system broke down twice in the home due to mechanical issues. The siblings liked Kiwi but the hardware issues influenced the reported scores. 
  
In other cases, participants were faced with the physical limitations of the system. For P7, as she grew more comfortable with Kiwi, she wanted to hug it, causing the face-plate covering the display to detach. The family requested that the robot in future iterations be sturdier so as to withstand physical affection from the child. Robot durability is a general challenge in any physical child-robot interactions, and can become an issue in SAR even if the interaction design does not call for physical interaction.

\section{CONCLUSION AND FUTURE WORK}
Overall, participants' responses to the SAR system were positive despite some technical constraints and challenges. The majority of participating families accepted the SAR system into their homes and reported the robot to be useful and likeable. Most child participants were also able to develop strong relationships with the robot over time. The hardware system, robot character, and interaction design can be improved to widen appeal and utility to a wider range of children with ASD. Having different themes for the games that children can choose from or another skin option for the robot can allow for personalization in the interaction and investment from a wider variety of participants. 

The research presented in this paper is part of our larger, ongoing research program toward long-term, in-home SAR for children with ASD. This paper specifically focused on the physical system, robot character, and interaction design of the fully autonomous SAR system. Our initial research findings encourage us to further investigate the social, emotional, and cognitive development of child participants over the course of intervention. We aim to develop and study SAR as a potential supplement to in-home therapy for children with ASD. This system is the first to be validated with a large cohort of children in their homes for month-long and longer use, providing insights and best practices for the designed and deployment of in-home, long-term HRI studies with children with ASD.

\addtolength{\textheight}{-4cm}   




\section*{ACKNOWLEDGMENT}

The authors thank David Becerra, Matthew Rueben, Isabel Rayas, and Thomas Groechel for their assistance and advice in preparing this paper. This research was supported by the NSF Expedition in Computing for Socially Assistive Robotics grant IIS-1139148.

\bibliographystyle{plain}
\bibliography{refs}

\begin{thebibliography}{10}

\bibitem{bainbridge2008effect}
Wilma~A Bainbridge, Justin Hart, Elizabeth~S Kim, and Brian Scassellati.
\newblock The effect of presence on human-robot interaction.
\newblock In {\em Robot and Human Interactive Communication, 2008. RO-MAN 2008.
  The 17th IEEE International Symposium on}, pages 701--706. IEEE, 2008.

\bibitem{baxter2013emergence}
Paul Baxter, James Kennedy, Tony Belpaeme, Rachel Wood, Ilaria Baroni, and
  Marco Nalin.
\newblock Emergence of turn-taking in unstructured child-robot social
  interactions.
\newblock In {\em Human-Robot Interaction (HRI), 2013 8th ACM/IEEE
  International Conference on}, pages 77--78. IEEE, 2013.

\bibitem{begum2016autism}
Momotaz Begum, Richard Serna, and Holly A.~Yanco.
\newblock Are robots ready to deliver autism interventions? a comprehensive
  review.
\newblock {\em International Journal of Social Robotics}, 8, 03 2016.

\bibitem{ilab_iser}
Balasubramanian Thiagarajan Zhonghao Shi-Leena Mathur Gisele~Ragusa
  Caitlyn~Clabaugh, Shomik~Jain and Maja Mataric.
\newblock Attentiveness of children with diverse needs to a socially assistive
  robot in the home.
\newblock In {\em International Symposium on Experimental Robotics (ISER)},
  2018.

\bibitem{christensen2016prevalence}
Deborah~L Christensen, Deborah~A Bilder, Walter Zahorodny, Sydney Pettygrove,
  Maureen~S Durkin, Robert~T Fitzgerald, Catherine Rice, Margaret
  Kurzius-Spencer, Jon Baio, and Marshalyn Yeargin-Allsopp.
\newblock Prevalence and characteristics of autism spectrum disorder among
  4-year-old children in the autism and developmental disabilities monitoring
  network.
\newblock {\em Journal of Developmental \& Behavioral Pediatrics}, 37(1):1--8,
  2016.

\bibitem{clabaugh2019escaping}
Caitlyn Clabaugh and Maja Matari\'{c}.
\newblock Escaping oz: Autonomy in socially assistive robotics.
\newblock {\em Annual Review of Control, Robotics, and Autonomous Systems},
  2(18):1--29, 2019.

\bibitem{clabaugh2015designing}
Caitlyn Clabaugh, Gisele Ragusa, Fei Sha, and Maja Matari\'{c}.
\newblock {Designing a socially assistive robot for personalized number
  concepts learning in preschool children}.
\newblock In {\em International Conference on Development and Learning and
  Epigenetic Robotics (ICDL-EpiRob)}, pages 314--319. IEEE, August 2015.

\bibitem{dautenhahn1999robots}
Kerstin Dautenhahn.
\newblock Robots as social actors: Aurora and the case of autism.
\newblock In {\em Proc. CT99, The Third International Cognitive Technology
  Conference, August, San Francisco}, volume 359, page 374, 1999.

\bibitem{dawson2008early}
Geraldine Dawson.
\newblock Early behavioral intervention, brain plasticity, and the prevention
  of autism spectrum disorder.
\newblock {\em Development and psychopathology}, 20(3):775--803, 2008.

\bibitem{diehl2012autism}
Joshua Diehl, Lauren Schmitt, Michael Villano, and Charles Crowell.
\newblock The clinical use of robots for individuals with autism spectrum
  disorders: A critical review.
\newblock {\em Research in autism spectrum disorders}, 6:249--262, 03 2012.

\bibitem{dover2007diagnose}
Clare~J Dover and Ann Le~Couteur.
\newblock How to diagnose autism.
\newblock {\em Archives of disease in childhood}, 92(6):540--545, 2007.

\bibitem{ekman1997face}
Paul Ekman and Erika~L Rosenberg.
\newblock {\em What the face reveals: Basic and applied studies of spontaneous
  expression using the Facial Action Coding System (FACS)}.
\newblock Oxford University Press, USA, 1997.

\bibitem{2017agencyASD}
David Feil-Seifer Maja J~Matari\'{c} Elaine Schaertl~Short, Eric C.~Deng.
\newblock Understanding agency in interactions between children with autism and
  socially assistive robots.
\newblock In {\em Journal of Human-Robot Interaction}, volume~6, pages 21--47,
  2017.

\bibitem{feilseifercomputational}
David Feil-Seifer and Maja Matari\'{c}.
\newblock Distance-based computational models for facilitating robot
  interaction with children.
\newblock {\em Journal of Human Robot Interaction}, 1, 07 2012.

\bibitem{fridin2014embodied}
Marina Fridin and Mark Belokopytov.
\newblock Embodied robot versus virtual agent: Involvement of preschool
  children in motor task performance.
\newblock {\em International Journal of Human-Computer Interaction},
  30(6):459--469, 2014.

\bibitem{gallese1998mirror}
Vittorio Gallese and Alvin Goldman.
\newblock Mirror neurons and the simulation theory of mind-reading.
\newblock {\em Trends in cognitive sciences}, 2(12):493--501, 1998.

\bibitem{gazzola2007anthropomorphic}
Valeria Gazzola, Giacomo Rizzolatti, Bruno Wicker, and Christian Keysers.
\newblock The anthropomorphic brain: the mirror neuron system responds to human
  and robotic actions.
\newblock {\em Neuroimage}, 35(4):1674--1684, 2007.

\bibitem{heerink2009measuring}
Marcel Heerink, Ben Krose, Vanessa Evers, and Bob Wielinga.
\newblock Measuring acceptance of an assistive social robot: a suggested
  toolkit.
\newblock In {\em Robot and Human Interactive Communication, 2009. RO-MAN 2009.
  The 18th IEEE International Symposium on}, pages 528--533. IEEE, 2009.

\bibitem{horner2005use}
Robert~H Horner, Edward~G Carr, James Halle, Gail McGee, Samuel Odom, and Mark
  Wolery.
\newblock The use of single-subject research to identify evidence-based
  practice in special education.
\newblock {\em Exceptional children}, 71(2):165--179, 2005.

\bibitem{kan2008autism}
CC~Kan, JK~Buitelaar, et~al.
\newblock Autism spectrum disorders in adults.
\newblock {\em Nederlands tijdschrift voor geneeskunde}, 152(24):1365--1369,
  2008.

\bibitem{kanne2008diagnostic}
Stephen~M Kanne, Jena~K Randolph, and Janet~E Farmer.
\newblock Diagnostic and assessment findings: A bridge to academic planning for
  children with autism spectrum disorders.
\newblock {\em Neuropsychology review}, 18(4):367--384, 2008.

\bibitem{lavelle2014economic}
Tara~A Lavelle, Milton~C Weinstein, Joseph~P Newhouse, Kerim Munir, Karen~A
  Kuhlthau, and Lisa~A Prosser.
\newblock Economic burden of childhood autism spectrum disorders.
\newblock {\em Pediatrics}, 133(3):e520--e529, 2014.

\bibitem{lee2006physically}
Kwan~Min Lee, Younbo Jung, Jaywoo Kim, and Sang~Ryong Kim.
\newblock Are physically embodied social agents better than disembodied social
  agents?: The effects of physical embodiment, tactile interaction, and
  people's loneliness in human--robot interaction.
\newblock {\em International Journal of Human-Computer Studies},
  64(10):962--973, 2006.

\bibitem{lee2005can}
Kwan~Min Lee, Namkee Park, and Hayeon Song.
\newblock Can a robot be perceived as a developing creature? effects of a
  robot's long-term cognitive developments on its social presence and people's
  social responses toward it.
\newblock {\em Human communication research}, 31(4):538--563, 2005.

\bibitem{leite2013social}
Iolanda Leite, Carlos Martinho, and Ana Paiva.
\newblock Social robots for long-term interaction: a survey.
\newblock {\em International Journal of Social Robotics}, 5(2):291--308, 2013.

\bibitem{leyzberg2012physical}
Daniel Leyzberg, Samuel Spaulding, Mariya Toneva, and Brian Scassellati.
\newblock The physical presence of a robot tutor increases cognitive learning
  gains.
\newblock In {\em Proceedings of the Cognitive Science Society}, volume~34,
  2012.

\bibitem{mataric2016socially}
Maja~J Matari{\'c} and Brian Scassellati.
\newblock Socially assistive robotics.
\newblock In {\em Springer Handbook of Robotics}, pages 1973--1994. Springer,
  2016.

\bibitem{moon2001extending}
Ji-Won Moon and Young-Gul Kim.
\newblock Extending the tam for a world-wide-web context.
\newblock {\em Information \& management}, 38(4):217--230, 2001.

\bibitem{ospina2008behavioural}
Maria~B Ospina, Jennifer~Krebs Seida, Brenda Clark, Mohammad Karkhaneh, Lisa
  Hartling, Lisa Tjosvold, Ben Vandermeer, and Veronica Smith.
\newblock Behavioural and developmental interventions for autism spectrum
  disorder: a clinical systematic review.
\newblock {\em PloS one}, 3(11):e3755, 2008.

\bibitem{rau2009effects}
PL~Patrick Rau, Ye~Li, and Dingjun Li.
\newblock Effects of communication style and culture on ability to accept
  recommendations from robots.
\newblock {\em Computers in Human Behavior}, 25(2):587--595, 2009.

\bibitem{robins2005robotic}
Ben Robins, Kerstin Dautenhahn, R~Te~Boekhorst, and Aude Billard.
\newblock Robotic assistants in therapy and education of children with autism:
  can a small humanoid robot help encourage social interaction skills?
\newblock {\em Universal Access in the Information Society}, 4(2):105--120,
  2005.

\bibitem{robins2008behaviour}
Ben Robins, Kerstin Dautenhahn, Ren{\'e} Te~Boekhorst, and Chrystopher~L
  Nehaniv.
\newblock Behaviour delay and robot expressiveness in child-robot interactions:
  a user study on interaction kinesics.
\newblock In {\em Proceedings of the 3rd ACM/IEEE international conference on
  Human robot interaction}, pages 17--24. ACM, 2008.

\bibitem{scassellati2012robots}
Brian Scassellati, Henny Admoni, and Maja Matari{\'c}.
\newblock Robots for use in autism research.
\newblock {\em Annual review of biomedical engineering}, 14:275--294, 2012.

\bibitem{scassellati2018improving}
Brian Scassellati, Laura Boccanfuso, Chien-Ming Huang, Marilena Mademtzi,
  Meiying Qin, Nicole Salomons, Pamela Ventola, and Frederick Shic.
\newblock Improving social skills in children with asd using a long-term,
  in-home social robot.
\newblock {\em Science Robotics}, 3(21):eaat7544, 2018.

\bibitem{short2017sprite}
Elaine Short, Dale Short, Yifeng Fu, and Maja~J Matari\'{c}.
\newblock Sprite: Stewart platform robot for interactive tabletop engagement.
  department of computer science, university of southern california.
\newblock {\em Tech Report}, 2017.

\bibitem{short2014train}
Elaine Short, Katelyn Swift-Spong, Jillian Greczek, Aditi Ramachandran,
  Alexandru Litoiu, Elena~Corina Grigore, David Feil-Seifer, Samuel Shuster,
  Jin~Joo Lee, Shaobo Huang, et~al.
\newblock How to train your dragonbot: Socially assistive robots for teaching
  children about nutrition through play.
\newblock In {\em Robot and Human Interactive Communication, 2014 RO-MAN: The
  23rd IEEE International Symposium on}, pages 924--929. IEEE, 2014.

\bibitem{singer2006play}
Dorothy~G Singer, Roberta~Michnick Golinkoff, and Kathy Hirsh-Pasek.
\newblock {\em Play= Learning: How play motivates and enhances children's
  cognitive and social-emotional growth}.
\newblock Oxford University Press, 2006.

\bibitem{van1992comparison}
Mary~E Van~Bourgondien, Lee~M Marcus, and Eric Schopler.
\newblock Comparison of dsm-iii-r and childhood autism rating scale diagnoses
  of autism.
\newblock {\em Journal of Autism and Developmental Disorders}, 22(4):493--506,
  1992.

\bibitem{wainer2006role}
Joshua Wainer, David~J Feil-Seifer, Dylan~A Shell, and Maja~J Matari\'{c}.
\newblock The role of physical embodiment in human-robot interaction.
\newblock In {\em Robot and Human Interactive Communication, 2006. ROMAN 2006.
  The 15th IEEE International Symposium on}, pages 117--122. IEEE, 2006.

\end{thebibliography}

\end{document}